\documentclass[letterpaper, 10 pt, conference]{ieeeconf}  

\IEEEoverridecommandlockouts                

\usepackage{graphicx}
\usepackage{float}
\usepackage{todonotes}
\usepackage{tabularx, booktabs}

\DeclareMathAlphabet\mathbfcal{OMS}{cmsy}{b}{n}

\def\etal{et~al.~}			  

\usepackage{amsmath}

\newcommand{\secref}[1]{Sec.~\ref{#1}}
\newcommand{\figref}[1]{Fig.~\ref{#1}} 
\newcommand{\tabref}[1]{Table~\ref{#1}}

\usepackage{amsmath}
\usepackage{subcaption}
\usepackage{amsfonts}
\usepackage{tikz}
\usepackage{multirow}
\newcommand{\tikzredcircle}[2][red,fill=red]{\tikz[baseline=-0.5ex]\draw[#1,radius=#2] (0,0) circle ;}
\newcommand{\tikzgreencircle}[2][green,fill=green]{\tikz[baseline=-0.5ex]\draw[#1,radius=#2] (0,0) circle ;}%
\usepackage[pagebackref=true,breaklinks=true,letterpaper=true,colorlinks,bookmarks=false]{hyperref}

\definecolor{amber}{rgb}{1.0, 0.75, 0.0}

\makeatletter
\let\NAT@parse\undefined
\makeatother
\usepackage[numbers]{natbib}

\title{\LARGE \bf NeRF-Supervision: Learning Dense Object Descriptors \\ from Neural Radiance Fields}

\author{
  \begin{tabular}{ccc}
  Lin Yen-Chen$^{1}$ \qquad & Pete Florence$^2$ \qquad & Jonathan T. Barron$^2$ \\  Tsung-Yi Lin$^{3*}$ \qquad & Alberto Rodriguez$^1$ \qquad & Phillip Isola$^1$
  \end{tabular}
  \vspace{0.1cm}
  \\
  $^{1}$MIT \qquad $^{2}$Google \qquad $^{3}$Nvidia
}

\begin{document}

\twocolumn[{%
\renewcommand\twocolumn[1][]{#1}%
\maketitle
\vspace{-1cm}

\begin{center}
		{
		\vspace{0.0cm}
		\normalsize
}
\centering

\url{https://yenchenlin.me/nerf-supervision/}
\vspace{0.25cm}

\href{https://www.youtube.com/watch?v=_zN-wVwPH1s}{
\includegraphics[width=\linewidth]{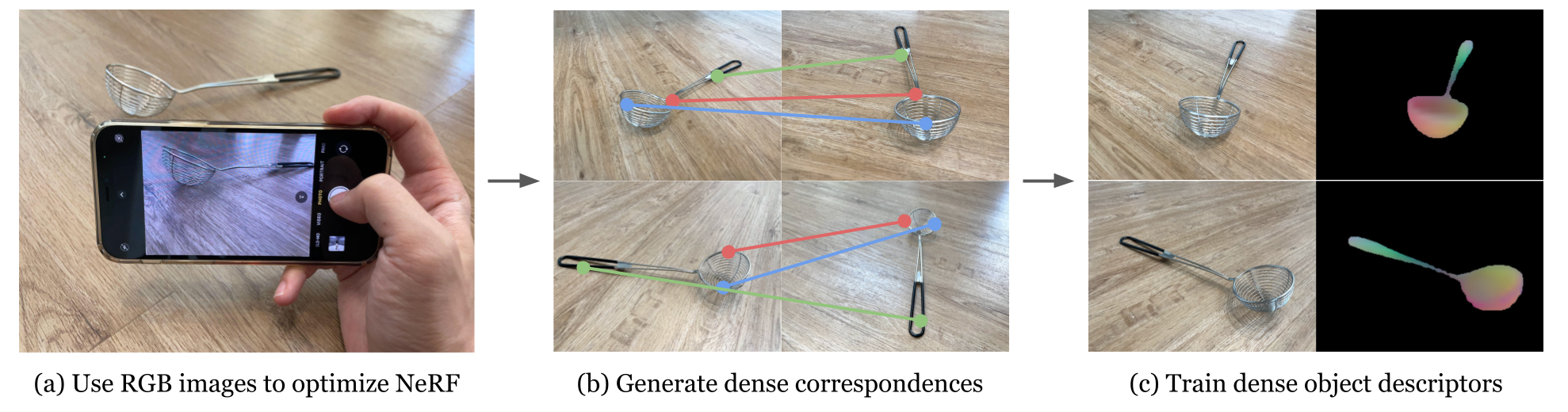}}
    \captionof{figure}{\textbf{Overview.} We present a new, RGB-sensor-only, self-supervised pipeline for learning object-centric dense descriptors, based on neural radiance fields (NeRFs)~\cite{mildenhall2020nerf}. The pipeline consists of three stages: (a) We collect RGB images of the object of interest and optimize a NeRF for that object; (b) The recovered NeRF's density field is then used to automatically generate a dataset of dense correspondences; (c) We use the generated dataset to train a model to estimate dense object descriptors, and evaluate that model on previously-unobserved real images. \textbf{Click the image to play the overview video in a browser.}}
    \label{fig:teaser}
\end{center}%
}]

{\let\thefootnote\relax\footnote{{$^*$Work done while at Google.}}}

\begin{abstract}
Thin, reflective objects such as forks and whisks are common in our daily lives, but they are particularly challenging for robot perception because it is hard to reconstruct them using commodity RGB-D cameras or multi-view stereo techniques.  
While traditional pipelines struggle with objects like these, Neural Radiance Fields (NeRFs) have recently been shown to be remarkably effective for performing view synthesis on objects with thin structures or reflective materials. In this paper we explore the use of NeRF as a new source of supervision for robust robot vision systems. In particular, we demonstrate that a NeRF representation of a scene can be used to train dense object descriptors. We use an optimized NeRF to extract dense correspondences between multiple views of an object, and then use these correspondences as training data for learning a view-invariant representation of the object. NeRF's usage of a density \emph{field} allows us to reformulate the correspondence problem with a novel distribution-of-depths formulation, as opposed to the conventional approach of using a depth map. Dense correspondence models supervised with our method significantly outperform off-the-shelf learned descriptors by 106\% (PCK@3px metric, more than doubling performance) and outperform our baseline supervised with multi-view stereo by 29\%. Furthermore, we demonstrate the learned dense descriptors enable robots to perform accurate 6-degree of freedom (6-DoF) pick and place of thin and reflective objects.

\end{abstract}

\section{Introduction}

Designing robust visual descriptors that are invariant to scale, illumination, and pose is a long-standing problem in computer vision~\cite{rublee2011orb, lowe2004distinctive, bay2008speeded}. Recently, learning-based visual descriptors, supervised by dense correspondences between images, have demonstrated superior performance compared to hand-crafted descriptors \cite{detone2018superpoint, GLUNet_Truong_2020, PDCNet_Truong_2021, jiang2021cotr}. However, producing the ground-truth dense correspondence data required for training these models is challenging, as the geometry of the scene and the poses of the cameras must somehow be estimated from an image (or known a priori). As a result, learning-based methods typically rely on either synthetically rendering an object from multiple views~\cite{dosovitskiy2015flownet,mayer2016large}, or on augmenting a non-synthetic image with random affine transformations from which ``ground truth'' correspondences can be obtained~\cite{GLUNet_Truong_2020,Rocco17,GOCor_Truong_2020}. While effective, these approaches have their limitations: the gap between real data and synthetic data may hinder performance, and data augmentation approaches may fail to identify correspondences involving out-of-plane rotation (which occur often in robot manipulation).

\begin{figure}
     \centering
     \begin{subfigure}[b]{0.23\textwidth}
         \centering
         \includegraphics[width=\textwidth]{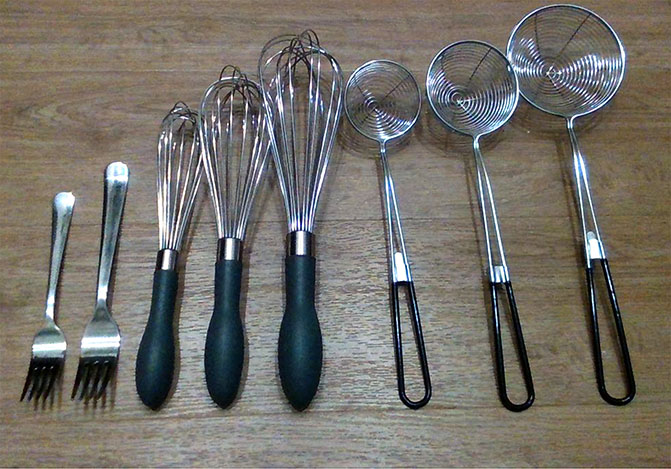}
         \caption{Objects}
         \label{fig:objects}
     \end{subfigure}
     \hfill
     \begin{subfigure}[b]{0.23\textwidth}
         \centering
         \includegraphics[width=\textwidth]{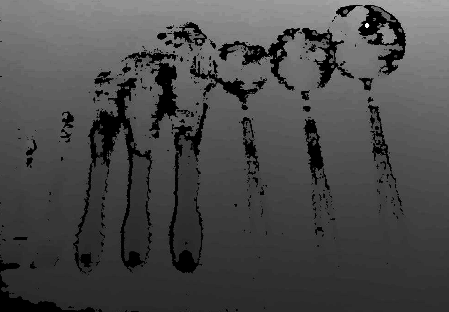}
         \caption{Depth camera image}
         \label{fig:realsense}
     \end{subfigure}
     \hfill
    \caption{\textbf{Motivation.} (a) Here we show the objects used in the work. Annotating dense correspondences for these objects is challenging because existing pipelines~\cite{florencemanuelli2018dense,manuelli2019kpam} rely on depth cameras, and therefore cannot capture thin or reflective objects. (b) This can be observed by visualizing the depth image from a commodity RGB-D camera (a RealSense D415), where the pixels colored black indicate where the depth sensor failed to produce a depth estimate.}
    \label{fig:motivation}
\end{figure}

\begin{figure}[t!]
    \centering
    \includegraphics[width=0.48\textwidth]{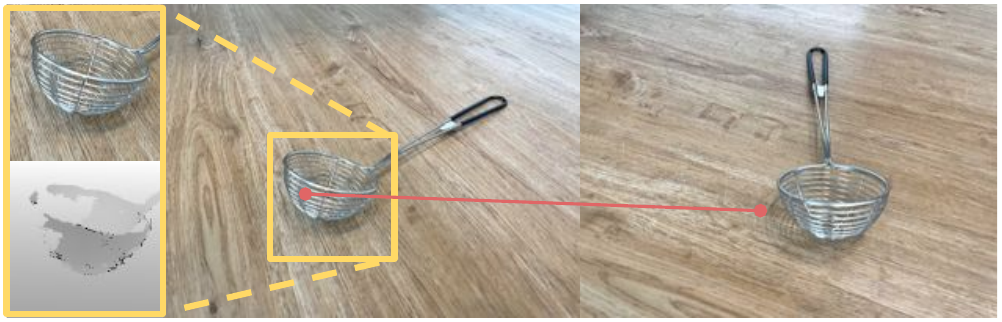}
    \caption{\textbf{Baselines.} Multi-view Stereo represents a potential alternative to depth cameras. However, the depth maps estimated by COLMAP (a widely-used MVS method)~\cite{schoenberger2016mvs} exhibit significant artifacts on thin or reflective objects, which leads to incorrect correspondences between pixels (shown in \textcolor{red}{red}).
    }
    \vspace{-0.4em}
    \label{fig:mvs}
\end{figure}

To learn a dense correspondence model, Florence \etal propose a self-supervised data collection approach based on robot motion in conjunction with a depth camera~\cite{florencemanuelli2018dense}. Their method generates dense correspondences given a set of posed RGB-D images and then uses them to supervise visual descriptors. 
However, this method works poorly for objects that contain thin structures or highly specular materials, as commodity depth cameras fail in these circumstances.
An object exhibiting thin structures or shiny reflectance, well-exemplified by objects such as forks and whisks, will result in a hole-riddled depth map (shown in ~\figref{fig:realsense}) which prevents the reprojection operation from generating high quality correspondences.
Multi-view stereo (MVS) methods present an alternative approach for solving this problem, as they do not rely on direct depth sensors and instead estimate depth using only RGB images.
However, conventional stereo techniques typically rely on patch-based photometric consistency, which implicitly assumes that the world is made of large and Lambertian objects. The performance of MVS is therefore limited in the presence of thin or shiny objects --- thin structures mean image patches may not reoccur across input images (as any patch will likely contain some part of the background, which may vary), and specularities mean that photometric consistency may be violated (as the object may look different when viewed from different angles).
Figure~\ref{fig:mvs} shows a failure case when applying COLMAP~\cite{schoenberger2016sfm}, a widely-used MVS method, on a strainer. Because COLMAP produces an incorrect depth map, the estimated correspondences are also incorrect.
To address the limitations of depth sensors and conventional stereo techniques, we introduce \textit{NeRF-Supervision} for learning object-centric dense correspondences: an RGB-only, self-supervised pipeline based on neural radiance fields (NeRF)~\cite{mildenhall2020nerf}. Unlike approaches based on RGB-D sensors or MVS, it can handle reflective objects as the view direction is taken as input for color prediction. Another advantage of using NeRF-Supervision over depth sensors or MVS is that the density field predicted by NeRF provides a mechanism for handling \emph{ambiguity} in photometric consistency: given a trained NeRF, the predicted density field can be used to sample a dataset of dense correspondences probabilistically. See~\figref{fig:teaser} for an overview of our method.
In our experiments, we consider 8 challenging objects (shown in~\figref{fig:objects}) and demonstrate that our pipeline can produce robust dense visual descriptors for all of them. Our approach significantly outperforms all off-the-shelf descriptors as well as our baseline method supervised with multi-view stereo. Furthermore, we demonstrate the learned dense descriptors enable robots to perform accurate 6-degree of freedom (6-DoF) pick and place of thin and reflective objects.

Our contributions are as follows: (i) a new, RGB-sensor-only, self-supervised pipeline for learning object-centric dense descriptors, based on neural radiance fields; (ii) a novel distribution-of-depths formulation, enabled by the estimated density field, which treats correspondence generation not via a single depth for each pixel, but rather via a distribution of depths; (iii) experiments showing that our pipeline can: (a) enable training accurate object-centric correspondences without depth sensors, and (b) succeed on thin, reflective objects on which depth sensors typically fail; and (iv) experiments showing that the distribution-of-depths formulation can improve the downstream precision of correspondence models trained on this data, when compared to the single-depth alternatives. 

\section{Related Work}
\noindent \textbf{Neural radiance fields.} 
NeRF is a powerful technique for novel view synthesis --- taking as input a set of images of an object, and producing novel views of that object~\cite{mildenhall2020nerf}. A central component of NeRF is the use of coordinate-based MLPs (neural networks that take as input a 3D coordinate in space) to estimate volumetric density and color in 3D. This MLP is embedded within a volumetric rendering engine, and gradient descent is used to optimize the weights of the scene to reproduce the input images, thereby resulting in an MLP that maps any input coordinate to a field of density (and color). Though NeRF has primarily been used for vision or graphics tasks such as appearance interpolation~\cite{martinbrualla2020nerfw} and portrait photography~\cite{park2021nerfies}, it has also been adopted for robotic applications such as pose estimation~\cite{yen2020inerf} and SLAM~\cite{SucarICCV2021}.
In this work, we propose using NeRF as a data generator for learning visual descriptors.

Note that NeRF represents \emph{all} scene content as a volumetric quantity --- everything is assumed to be some degree of semi-transparent, and ``hard'' surfaces are simulated using a very dense (but not infinitely dense) field~\cite{drebin1988volume}. Though the use of volumetric rendering provides significant benefits (most notably, smooth gradient-based optimization) it does present some difficulties when attempting to use NeRF in a robotics context, as NeRF does not directly estimate the boundaries of objects nor does it directly produce depth maps. However, the density field estimated by NeRF can be used to synthesize depth maps by computing the \emph{expected termination depth} of a ray --- a ray is cast towards the camera, and the  density field is used to determine how "deep" into the volumetric object that ray is expected to penetrate, and that distance is then used as a depth map~\cite{mildenhall2020nerf}. Some recent work has explored improving these depth maps, such as Deng \etal \cite{kangle2021dsnerf} who use the depths estimated by COLMAP to directly supervise these depth maps.

\noindent \textbf{Dense descriptors.} \indent 
Dense visual descriptors play an important role in 3D scene reconstruction, localization, object pose estimation, and robot manipulation \cite{florencemanuelli2018dense, schmidt2016self, florence2019self, manuelli2020keypoints, sundaresan2020learning, ganapathi2020learning}.  Modern approaches rely on machine learning to learn a visual descriptor: First, image pairs with annotated correspondences are obtained, either by a generative approach or through manual labeling. Then these correspondences are used as training data to learn pixel-level descriptors such that the feature embeddings of corresponding pixels are similar. 
A common approach for generating data is to use synthetic warping with large image collections, as is done by GLU-Net~\cite{GLUNet_Truong_2020}. Despite the benefit of being trained with many examples, these methods often fail to predict correspondences in images that exhibit out-of-plane rotation, as image-space warping only demonstrates in-plane rotation. Other approaches leverage explicit 3D geometry to supervise correspondences \cite{schmidt2016self,choy2016universal}. Within this category, Florence \etal~\cite{florencemanuelli2018dense} demonstrate a self-supervised learning approach for collecting training correspondences using motion and depth sensors on robots. This approach is prone to failure whenever the depth sensors fails to measure the correct depth, which occurs often for thin or reflective structures. Methods that use only RGB inputs face the challenge of ambiguity of visual correspondences on regions with no texture or drastic depth variations. Other approaches have demonstrated simulation-based descriptor training \cite{sundaresan2020learning,ganapathi2020learning}, which is an attractive approach due to its flexibility. However, it requires significant engineering effort to configure accurate and realistic simulations. Our work uses NeRF to generate training correspondences from only real-world non-synthetic RGB images captured in uncontrolled settings, thereby avoiding the shortcomings of depth sensors and addressing ambiguity by modeling correspondence with a density field, which we interpret as a probability distribution over possible depths.

\section{Method}

Our approach introduces an RGB-sensor-only framework to provide training data for supervising dense correspondence models. In particular, the framework provides the fundamental unit of training data required for training such models, which is a tuple of the form: 
\begin{equation}
(\mathbf{I_s}, \mathbf{u}_s, \mathbf{I}_t, \mathbf{u}_t) 
\label{eq:training-tuples}
\end{equation}
that consists of a pair of RGB images $\mathbf{I}_s$ and $\mathbf{I}_t$, each  in $\mathbb{R}^{w\times h \times 3}$, and a pair of pixel-space coordinates $\mathbf{u}_s$ and $\mathbf{u}_t$, each in $\mathbb{R}^2$, whose image-forming rays intersect the same point in 3D space.  Rather than proposing a specific correspondence model for using these tuples, our focus is on an approach for {\em{generating}} this training data. 

Given this ground-truth correspondence data (\ref{eq:training-tuples}), a variety of  learning-based correspondence approaches can be trained, but our experiments focus on {\em{object-centric}} dense descriptor models \cite{florencemanuelli2018dense} which have been shown to be useful in enabling generalizable robot manipulation \cite{florencemanuelli2018dense, florence2019self, manuelli2020keypoints, sundaresan2020learning, ganapathi2020learning}.
With a descriptor-based correspondence model, a neural network $f_{\theta}$ with parameters $\theta$ maps an input RGB image $\mathbf{I}$ to a dense visual descriptor image $f_{\theta}(\mathbf{I}) \in \mathbb{R}^{h \times w \times d}$ where each pixel is encoded by a $d$-dimensional feature vector, and closeness (small Euclidean distance) in the descriptor space indicates correspondence despite viewpoint changes, lighting changes, and potentially category-level variation \cite{florencemanuelli2018dense, schmidt2016self, choy2016universal}.

\subsection{NeRF Preliminaries} \label{sec:nerf_pre}
NeRF \cite{mildenhall2020nerf} use a neural network to represent a scene as a volumetric field of density $\sigma$ and RGB color $\mathbf{c}$. The weights of a NeRF are initialized randomly and optimized for an individual scene using a collection of input RGB images as supervision (the camera poses of the images are assumed to be known, and are often recovered via COLMAP\cite{schoenberger2016sfm}). After optimization, the density field modeled by the NeRF captures the geometry of the scene (where a large density indicates an occupied region) and the color field models the view-dependent appearance of those occupied regions. 
A multilayer perceptron (MLP) parameterized by weights $\Theta$ is used to predict the density $\sigma$ and RGB color $\mathbf{c}$ of each point as a function of that point's 3D position $\mathbf{x} = (x, y, z)$ and unit-norm viewing direction $\mathbf{d}$ as input. To overcome the spectral bias that neural networks exhibit in low dimensional spaces~\cite{tancik2020fourfeat}, each input is encoded using a positional encoding $\gamma(\cdot)$, giving us $(\sigma, \mathbf{c}) \leftarrow F_{\Theta}(\gamma(\mathbf{x}), \gamma(\mathbf{d}))$.
To render a pixel, NeRF casts a camera ray $\mathbf{r}(t) = \mathbf{o} + t\mathbf{d}$ from the camera center $\mathbf{o}$ along the direction $\mathbf{d}$ passing through that pixel on the image plane. Along the ray, $K$ discrete points $\{\mathbf{x}_k = \mathbf{r}(t_k)\}_{k=1}^K$ are sampled for use as input to the MLP, which outputs a set of densities and colors $\{\sigma_k, \mathbf{c}_k\}_{k=1}^K$. These values are then used to estimate the color $\hat{\textbf{C}}(r)$ of that pixel following volume rendering~\cite{kajiya84}, using a numerical quadrature approximation~\cite{max95}:
\newcommand{\bigT}{T}
\begin{equation} \label{eq:volume_rendering}
\begin{split}
    \hat{\textbf{C}}(\mathbf{r}) = \sum_{k=1}^{K} \bigT_k \bigg(1- \exp\Big(-\sigma_k (t_{k+1} - t_k)\Big)\bigg) \textbf{c}_k, \\ \text{with} \quad \bigT_k = \text{exp} \Big(-\sum_{k' < k} \sigma_{k'} (t_{k'+1} - t_{k'})\Big)
\end{split}
\end{equation}
where $\bigT_k$ can be interpreted as the probability that the ray successfully transmits to point $\mathbf{r}(t_k)$.
NeRF is then trained to minimize a photometric loss $\mathcal{L_{\text{photo}}} = \sum_{\mathbf{r} \in \mathbfcal{R}} ||\hat{\mathbf{C}}(\mathbf{r}) - \mathbf{C}(\mathbf{r})||_2^2$,
using some sampled set of rays $\mathbf{r} \in \mathbfcal{R}$ where $\mathbf{C}(\mathbf{r})$ is the observed RGB value of the pixel corresponding to ray $\mathbf{r}$ in some image. 
%
For more details, we refer readers to Mildenhall \etal \cite{mildenhall2020nerf}.

\begin{figure*}[t!]
     \centering
     \begin{subfigure}[b]{0.48\textwidth}
         \centering
         \includegraphics[width=\textwidth]{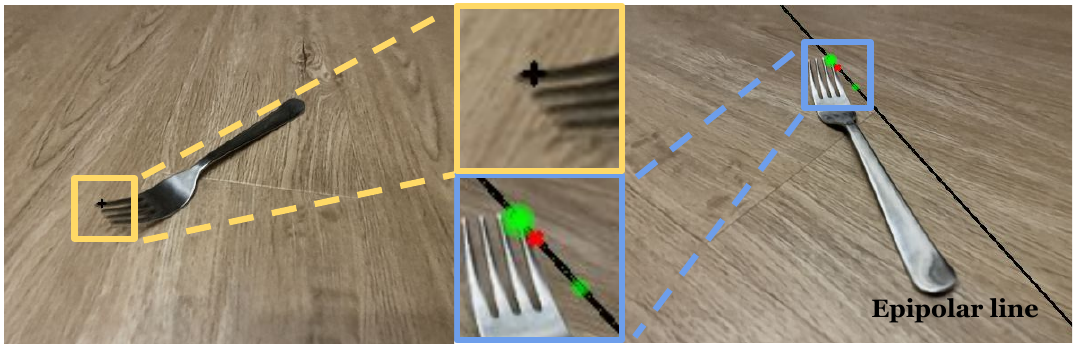}
         \caption{Fork}
         \label{fig:fork_mean_depth}
     \end{subfigure}
     \hfill
     \begin{subfigure}[b]{0.48\textwidth}
         \centering
         \includegraphics[width=\textwidth]{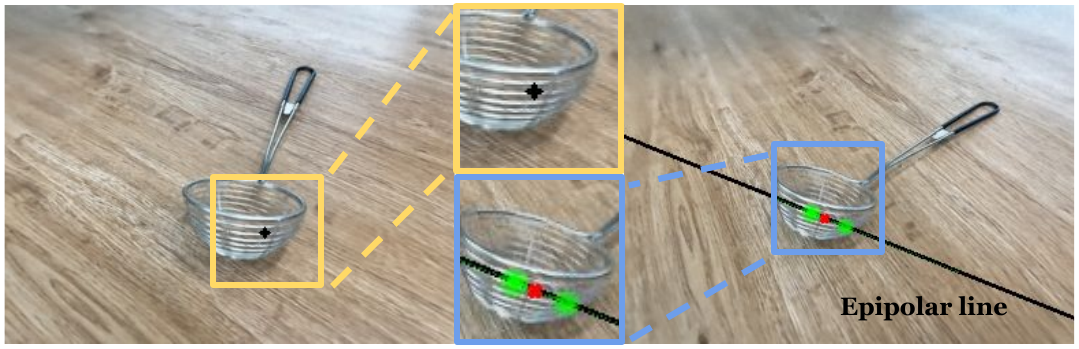}
         \caption{Strainer}
         \label{fig:strainer_mean_depth}
     \end{subfigure}
     \hfill
    \caption{\textbf{Generating correspondences from NeRF's density field vs. depth map}. We denote the query pixel $\mathbf{u}_s$ as \textbf{+}, the correspondence found in the other image using NeRF's depth map as \tikzredcircle{2pt}, and correspondences found by NeRF's density field as \tikzgreencircle{2pt}, where each point's radius is scaled by its corresponding weight. We show two example objects: (a) fork and (b) strainer. The correspondence implied by NeRF's depth map is incorrect, but by using NeRF's density field directly, the correct correspondence can be sampled probabilistically.}
    \label{fig:mean_depth_is_wrong}
    \vspace{-0.4em}
\end{figure*}

\subsection{Sparse Depth Supervision for NeRF}\label{sec:depth-supervision}
For objects and scenes with particularly challenging geometry (in particular, thin and reflective structures), we find that leveraging recent work on incorporating depth supervision into NeRF \cite{kangle2021dsnerf} improves geometry accuracy for our purposes. Though Deng \etal \cite{kangle2021dsnerf} focus on the few-image setting (i.e. $\sim\!5$ images), in our investigations we found that even in the many-view (i.e. $\sim\!60$ images) setting, adding depth supervision is beneficial. 
Specifically, we find NeRF's density prediction often deteriorates in real-world 360$^{\circ}$ inward-facing scenes due to the transient shadows cast by the photographer or robot on the scene. Because these shadows appear in some images but not others, NeRF tends to explain them away by introducing artifacts in the optimized density field. Incorporating the depth supervision appears to effectively mitigate this issue. 

Though NeRF's primary goal is to perform view synthesis by rendering RGB images, the volumetric rendering equation in \eqref{eq:volume_rendering} can be modified slightly to produce the expected termination depth of each ray (as was done in \cite{mildenhall2020nerf, kangle2021dsnerf}) by simply replacing the predicted color $\textbf{c}_k$ with the distance $t_k$:
\begin{equation} \label{eq:volume_rendering_depth}
\begin{split}
    \hat{\mathbf{D}}(\mathbf{r}) = \sum_{k=1}^{K} \bigT_k \bigg(1- \exp\Big(-\sigma_k (t_{k+1} - t_k)\Big)\bigg) t_k\,. 
\end{split}
\end{equation}
\newcommand{\keypoint}{k} 
Because $\bigT_k$ represents the probability of the ray transmitting through interval $k$, the resulting depth $\hat{\mathbf{D}}(\mathbf{r})$ is the expected distance that ray $\mathbf{r}$ will travel when cast into the scene.
We can obtain a ground-truth depth $\mathbf{D}(\mathbf{r})$ by first transforming the 3D keypoint $\mathbf{\keypoint}(\mathbf{r})$ that is associated with the ray $\mathbf{r}$ to the camera frame with camera pose $\mathbf{G} \in \text{SE(3)}$ and then extract its coordinate along the camera's $z$-axis:
$
\mathbf{D}(\mathbf{r}) = \langle \mathbf{G}^{-1} \mathbf{\keypoint}(\mathbf{r}), [0, 0, 1] \rangle 
$.
The depth-supervision loss $\mathcal{L}_{\text{depth}} = \sum_{\mathbf{r} \in \mathbfcal{R}} \lVert \hat{\mathbf{D}}(\mathbf{r}) - \mathbf{D}(\mathbf{r}) \rVert_2^{2}$ is defined as the squared distance between the predicted depth $\hat{\mathbf{D}}(\mathbf{r})$ and the ``ground-truth'' depth $\mathbf{D}(\mathbf{r})$ (which in our case is the partial depth map generated by COLMAP's structure from motion). Note this supervision is only sparse, not dense --- this loss is not imposed for pixels where the depth supervisor does not return a valid depth. The final combined loss for training DS-NeRFs is: $\mathcal{L} = \mathcal{L}_{\text{photo}} + \mathcal{L}_{\text{depth}}$.

\subsection{Depth-Map Dense Correspondences from NeRF}\label{sec:mean-depth-correspondences}

The first approach we investigate in order to generate correspondence training data from NeRF is to {\em{render pairs of RGB-D images}}, and effectively treat NeRF as a traditional depth sensor by extracting a {\em{depth-map}} $\mathbf{D}\in \mathbb{R}^{w \times h}$ with a single-valued depth at each discrete pixel.  In this case, the single-valued depth estimate for each dense pixel is computed using \eqref{eq:volume_rendering_depth}.  Each training image pair consists of one rendered RGB-D image $(\hat{\mathbf{I}}_s, \hat{\mathbf{D}}_s)$ with camera pose $\mathbf{G}_s$ and another rendered RGB-D image $(\hat{\mathbf{I}}_t, \hat{\mathbf{D}}_t)$ with camera pose $\mathbf{G}_t$. Below, we slightly abuse the notation and use $\hat{\mathbf{D}}_s(\mathbf{u}_s)$ to represent the predicted depth at pixel $\mathbf{u}_s$.

Given these depth maps rendered by NeRF, and assuming known camera intrinsics $\mathbf{K}$, we can then generate the target pixel $\mathbf{u}_t$ in $\hat{\mathbf{I}}_t$ given a query pixel $\mathbf{u}_s$ in $\hat{\mathbf{I}}_s$:
\begin{equation} \label{eq:projection}
\begin{split}
    \mathbf{u}_t = \pi\Big( \mathbf{K} {\mathbf{G}_t}^{-1} \mathbf{G_s} \mathbf{K}^{-1} \hat{\mathbf{D}}_s(\mathbf{u}_s) \mathbf{u}_s \Big)
\end{split}
\end{equation}
where $\pi(\cdot)$ represents the projection operation. 
We will refer to this data generation method as \textit{depth-map}, as it uses the mean of NeRF's distribution of depths at each pixel to render a depth map.

\subsection{Generating Probabilistic Dense Correspondences from NeRF's Density Field}\label{sec:depth-field-correspondences}

While using NeRF's depth map to generate dense correspondences may work well when the distribution of density along the ray has only a single mode, it may produce an incorrect depth when the density distribution is multi-modal along the ray.  In~\figref{fig:mean_depth_is_wrong}, we show two examples of this case, where NeRF's depth map generates incorrect correspondences.
To resolve this issue, we propose to treat  correspondence generation not via a single depth for each pixel, but via a  distribution of depths, which as shown in  ~\figref{fig:mean_depth_is_wrong} can have modes which correctly recover correspondences where the depth map failed.

Specifically, we can sample depth values based on the alpha compositing weights $w$:
\begin{equation} \label{eq:depth_probability}
\begin{split}
    w(\hat{\mathbf{D}}(\mathbf{u}_s) = t_k) = \bigT_k \bigg(1- \exp\Big(-\sigma_k (t_{k+1} - t_k)\Big)\bigg)
\end{split}
\end{equation}

Rather than reducing the depth distribution into its mean by rendering out depth maps and sampling the correspondences deterministically, this formulation {\em{retains a complete distribution over depths}} and samples correspondences probabilistically.  In practice, we first sample $K$ points along each ray and get $\{w(\hat{\mathbf{D}}(\mathbf{u}_s) = t_k), t_k\}_{k=1}^K$ from NeRF. Then, we normalize $\{w(\hat{\mathbf{D}}(\mathbf{u}_s) = t_k)\}_{k=1}^K$ to sum to 1 and treat it as a probability distribution for sampling $t$.

We hypothesize the probabilistic formulation can produce more precise downstream neural correspondence networks, since as depicted in Fig.~\ref{fig:mean_depth_is_wrong}, the {\em{modes}} of the density, rather than the {\em{mean}}, can be closer to the ground truth. Furthermore, when combined with a self-consistency check (\secref{sec:check}) during descriptor learning, the probability of sampling false positives is reduced.
This hypothesis is tested in our Results section.
 
\subsection{Additional Correspondence Learning Details}\label{sec:check}

\noindent \textbf{Self-consistency.} \indent After obtaining $\mathbf{u}_t$ from $\mathbf{u}_s$, we perform a self-consistency check by starting from $\mathbf{u}_t$ and identify its probabilistic correspondence $\hat{\mathbf{u}}_s$ in $I_s$. We only adopt the pair of pixels $(\mathbf{u}_s, \mathbf{u}_t)$ if the distance between $\mathbf{u}_s$ and $\hat{\mathbf{u}}_s$ is smaller than certain threshold. This is our probabilistic analogue to the deterministic visibility check in \cite{florencemanuelli2018dense,trucco1998introductory}.

\noindent \textbf{Sampling from mask.} \indent We acquire object masks for the training images through a finetuned Mask R-CNN~\cite{he2017mask}. Similar to Dense Object Nets~\cite{florencemanuelli2018dense}, masks are used to sample pixels of the object during descriptor learning. 

\section{Results}
We execute a series of experiments using real world images for training and evaluation. We evaluate dense descriptors learned with correspondences generated with different approaches.
The goals of the experiments are four-fold: (i) to investigate whether the 3D geometry predicted by NeRF is sufficient for training precise descriptors, particularly on challenging thin and reflective objects, (ii) to compare our proposed method to existing off-the-shelf descriptors, (iii) to investigate whether the distribution-of-depth formulation is effective, and (iv) to test the generalization ability of visual descriptors produced by our pipeline.

\subsection{Settings}
\noindent \textbf{Datasets.} \indent We evaluate our approach and baseline methods using 8 objects (3 distinct classes). For each object, we captured 60 input images using an iPhone 12 with locked auto exposure and auto focus. The images are resized to $504 \times 378$. We use COLMAP~\cite{schoenberger2016sfm} to estimate both camera poses and sparse point cloud of each object. To construct the test set, 8 images are randomly selected and held-out during training. We manually annotate (for evaluation only) 100 correspondences using these test images for each object.

\noindent \textbf{Metrics.} We employ the Average End Point Error (AEPE) and Percentage of Correct Keypoints (PCK) as the evaluation metrics. 
AEPE is computed as the average Euclidean distance, in pixel space, between estimated and ground-truth correspondences.
PCK@$\delta$ is defined as the percentage of estimated correspondences with a pixel-wise Euclidean distance $<\delta$  w.r.t. to the ground-truths.

\subsection{Methods}
First, we consider several off-the-shelf learned descriptors that attain state-of-the-art results on commonly used dense correspondence benchmarks (e.g., ETH3D~\cite{schops2017multi}).

\begin{itemize}
    \item \textbf{GLU-Net}~\cite{GLUNet_Truong_2020} is a model architecture that integrates both global and local correlation in a feature pyramid-based network for estimating dense correspondences. 
    
    \item \noindent \textbf{GOCor}~\cite{GOCor_Truong_2020} improves GLU-Net~\cite{GLUNet_Truong_2020}'s feature correlation layers to disambiguate similar regions in the scene.
    
    \item \textbf{PDC-Net}~\cite{PDCNet_Truong_2021} adopts the architecture of GOCor~\cite{GOCor_Truong_2020} and further parametrizes the predictive distribution as a constrained mixture model for estimating dense correspondences and their uncertainties.
\end{itemize}

Next, we train Dense Object Nets (DONs)~\cite{florencemanuelli2018dense} for learning dense visual descriptors. In practice, we set the dimensionality of visual descriptors $d=3$. We consider using COLMAP or NeRF to generate training correspondences to supervise DONs.
\begin{itemize}
\item \textbf{COLMAP}~\cite{schoenberger2016mvs} is a widely-used classical Multi-view Stereo (MVS) method. We use the estimated depth maps to generate correspondences.
\item \textbf{NeRF}~\cite{mildenhall2020nerf} is a volume-rendering approach, which we either use via depth maps (Sec.~\ref{sec:mean-depth-correspondences})  or probabilistically through the density field (Sec.~\ref{sec:depth-field-correspondences}) to generate correspondences.
\end{itemize}

\subsection{Comparisons} 
We evaluate dense descriptors and show quantitative results in~\tabref{table:AEPE}, ~\tabref{table:PCK3}, and ~\tabref{table:PCK5}.
We find the off-the-shelf dense descriptors do not work well to handle object-centric scenes, potentially because they are trained on images with synthetic warp and have not seen the target objects from a wide range of viewing angles.
In contrast, Dense Object Nets trained with target objects perform much better. This suggests the need of a data collection pipeline to generate object-centric training data for robot manipulation. Among the three correspondence generation approaches, COLMAP has the highest error compared to other methods. Using the density field of NeRF to sample correspondences attains the best performance. It outperforms Dense Object Nets with COLMAP by 29\% and off-the-shelf descriptors by 106\% on PCK@3px  metric.

\begin{figure*}[t!]
     \centering
     \begin{subfigure}[b]{0.32\textwidth}
         \centering
         \includegraphics[width=\textwidth]{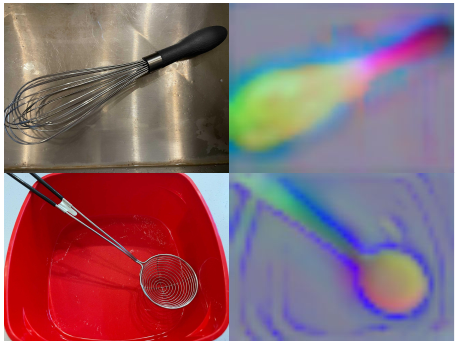}
         \caption{Different background and shadow}
         \label{fig:results_background}
     \end{subfigure}
     \hfill
     \begin{subfigure}[b]{0.32\textwidth}
         \centering
         \includegraphics[width=\textwidth]{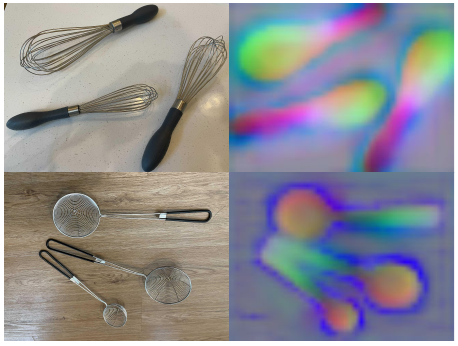}
         \caption{Multiple objects}
         \label{fig:results_multi}
     \end{subfigure}
     \hfill
     \begin{subfigure}[b]{0.32\textwidth}
         \centering
         \includegraphics[width=\textwidth]{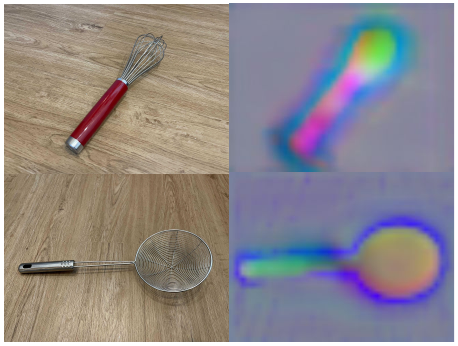}
         \caption{Unseen objects}
         \label{fig:results_unseen}
     \end{subfigure}
    \caption{\textbf{Qualitative results of generalization to novel scenes and objects.} (a) We show the learned object descriptors can be consistent across significant 1)  viewpoint, 2) background, and 3) lighting variations. (b) We visualize the learned descriptors for multiple objects despite the models have never seen multiple objects during training. (c) We test our model on objects that are not seen during training. The visual descriptors are shown to be consistent with previously-seen objects in the category.}
    \label{fig:results}
\end{figure*}

\vspace{-0.05cm}
\subsection{Generalization}
We evaluate the trained Dense Object Nets on novel scenes and objects not present in the training data. \figref{fig:results} shows examples of Whisks and Strainers and their visual descriptors. We follow the same visualization method in~\cite{florencemanuelli2018dense}.

\noindent \textbf{Noisy background and lighting.} In~\figref{fig:results_background}, we show results of our learned descriptors when the objects are placed on a different background or in different lighting conditions. The results demonstrate that our learned descriptors can be deployed in environments different from the training scenes.

\noindent \textbf{Multiple objects.} We show the learned descriptors when the input image contains multiple objects in~\figref{fig:results_multi}. The results demonstrate that the descriptors are consistent for objects of different sizes.

\noindent \textbf{Category-level generalization.}
We further test our model on unseen objects of the same category. \figref{fig:results_unseen} shows unseen objects not in the training set. The learned visual descriptors can robustly generalize to these unseen objects and estimate the view-invariant descriptors.

\begin{table*}[t!]
\begin{center}
\caption{Average End Point Error (AEPE),  	$\downarrow$ lower is better.}
\vspace{-0.15cm}
{\renewcommand{\arraystretch}{1.4}
\label{table:AEPE}
\begin{tabular*}{\textwidth}{ll @{\extracolsep{\fill}} cccccccc | c}
\toprule
    && Strainer-S & Strainer-M & Strainer-L & Whisk-S & Whisk-M & Whisk-L & Fork-S & Fork-L & Mean
\\ \midrule
\multirow{3}{*}{\em{Off-the-shelf}} & GLU-Net~\cite{GLUNet_Truong_2020}   &   33.25   &   28.09   &   28.92   &   16.06   &   15.36   &   39.04 &     17.12   &	18.28   &	24.52
\\
& GOCor~\cite{GOCor_Truong_2020}  &   34.23	&   26.89   &	20.92   &	10.8    &	7.04    &	31.95 & 10.2    &    13.86   &	19.49
\\
& PDC-Net~\cite{PDCNet_Truong_2021}   &   32.48   &	13.7    &	23.77   &	7.82    &	5.81    &	19.94 & 8.3  &	8.76    &	15.07
\\ \midrule
\multirow{3}{*}{\em{DON\cite{florencemanuelli2018dense} via}} & Depth map, COLMAP MVS      & 8.91   &   5.52    &	7.65    &   4.50 &   4.10    &   8.90 &     5.31    &   5.87    &	6.35
\\
& Depth map, NeRF (ours) &   5.64	&   4.31	&   5.24	&   3.82	&   3.52	&   6.84 &  3.73    &	4.19    &	4.66
\\
& Density field, NeRF (ours)   &   \textbf{4.53}    &	\textbf{4.08}    &	\textbf{3.93}    &	\textbf{3.28}    &	\textbf{3.19}    &	\textbf{4.96} &  \textbf{3.42}    &   \textbf{3.66}    &   \textbf{3.88}
\\ \bottomrule
\end{tabular*}
}
\vspace{-0.2cm}
\end{center}
\end{table*}

\begin{table*}[t!]
\begin{center}
\caption{Percentage Correct Keypoints (PCK@3px) for 3 pixels, $\uparrow$ higher is better.}
\vspace{-0.15cm}
{\renewcommand{\arraystretch}{1.4}
\label{table:PCK3}
\begin{tabular*}{\textwidth}{ll @{\extracolsep{\fill}} cccccccc | c}
\toprule
    && Strainer-S & Strainer-M & Strainer-L & Whisk-S & Whisk-M & Whisk-L & Fork-S & Fork-L & Mean
\\ \midrule
\multirow{3}{*}{\em{Off-the-shelf}} & GLU-Net~\cite{GLUNet_Truong_2020}   &   0.04    &	0.04    &	0.07    &	0.24    &	0.26    &	0   &  0.16    &	0.14    &	0.12
\\
& GOCor~\cite{GOCor_Truong_2020}  & 0.1   &	0.05    &	0.07    &	0.26    &	0.33    &	0.03    &   0.18   &   0.16    &   0.15
\\
& PDC-Net~\cite{PDCNet_Truong_2021}   &   0.14    &	0.19    &	0.11    &	0.48    &	0.51    &	0.19 & 0.42	&   0.38    &	0.30
\\ \midrule
\multirow{3}{*}{\em{DON\cite{florencemanuelli2018dense} via}} & Depth map, COLMAP MVS   &   0.32    &	0.41    &	0.38    &	0.57    &	0.64    &	0.44    &  0.55 &   0.51    &   0.48
\\
& Depth map, NeRF (ours)    &   0.52    &	0.56    &	0.51    &	0.62    &	\textbf{0.66}    &	0.50 & 0.67  &	0.63    &   0.58
\\
& Density field, NeRF (ours)   &   \textbf{0.58}    &	\textbf{0.59}    &	\textbf{0.61}    &	\textbf{0.64}    &	\textbf{0.66}    &	\textbf{0.58}    &   \textbf{0.69}   &   \textbf{0.64}    &   \textbf{0.62}
\\ \bottomrule
\end{tabular*}
}
\vspace{-0.2cm}
\end{center}
\end{table*}

\begin{table*}[t!]
\begin{center}
\caption{Percentage Correct Keypoints (PCK@5px) for 5 pixels, $\uparrow$ higher is better.}
\vspace{-0.15cm}
{\renewcommand{\arraystretch}{1.4}
\label{table:PCK5}
\begin{tabular*}{\textwidth}{ll @{\extracolsep{\fill}} cccccccc | c}
\toprule
    & & Strainer-S & Strainer-M & Strainer-L & Whisk-S & Whisk-M & Whisk-L & Fork-S & Fork-L & Mean
\\ \midrule

\multirow{3}{*}{\em{Off-the-shelf}} & GLU-Net~\cite{GLUNet_Truong_2020}   &   0.09    &	0.09&	0.10&	0.37&	0.44&	0.06&	0.26&	0.21&	0.20
\\
& GOCor~\cite{GOCor_Truong_2020}  & 0.13& 	0.1&	0.11&	0.47&	0.63&	0.09&	0.29&	0.28&	0.26
\\
& PDC-Net~\cite{PDCNet_Truong_2021}   &   0.29&	0.25&	0.16&	0.53&	0.68&	0.26&	0.57&	0.51&	0.41
\\ \midrule
\multirow{3}{*}{\em{DON\cite{florencemanuelli2018dense} via}} & Depth map, COLMAP MVS   &   0.62    &	0.72&	0.64&	0.79&	0.80&	0.48&	0.60&	0.55&	0.65
\\
& Depth map, NeRF (ours)   &   0.82   &	0.84    &	0.75    &	\textbf{0.82}    &	0.81    &	0.56    &	0.79    &	0.76    &	0.77
\\
& Density field, NeRF (ours)   &    \textbf{0.84}    &	\textbf{0.87}    &	\textbf{0.79}    &	\textbf{0.82}    &	\textbf{0.82}    &	\textbf{0.64}    &	\textbf{0.82}    &	\textbf{0.78}    &	\textbf{0.80}
\\ \bottomrule
\end{tabular*}
}
\vspace{-0.2cm}
\end{center}
\end{table*}

\subsection{Example Application: 6-DoF Robotic Pick and Place}
We demonstrate accurate 6-DoF pick and place of thin and reflective objects. After learning the dense descriptors, we specify a set of semantic keypoints which encode a SE(3) grasp pose for each category. Before any grasp, we track keypoints' 2D locations using the descriptors and move the robot to capture two RGB images of the scene using the camera mounted on the robot arm. Then, we use triangulation to derive keypoints' 3D locations and execute the encoded SE(3) grasp pose. For more details, please see Sec.~\ref{sec:robot_details}.

\section{Conclusion}
We introduce NeRF-Supervision as a pipeline to generate data for learning object-centric dense descriptors. 
Compared to previous approaches based on RGB-D cameras or MVS, our method enables learning dense descriptors of thin, reflective objects.
We believe these results chart forward a general paradigm in which NeRF may be leveraged as an untapped representational format for supervising robot vision systems.

\vspace{5mm}
\noindent{\textbf{Acknowledgements.}} We thank Felix Yanwei Wang, Anthony Simeonov, Wei-Chiu Ma, Rachel Holladay, and Maria Bauza for helpful feedback on the draft. This work was supported by a grant from Amazon.

\begin{appendix}
\subsection{Robotic Pick And Place}\label{sec:robot_details}
\begin{figure}[h!]
    \centering
    \href{https://youtu.be/_zN-wVwPH1s?t=238}{
    \begin{subfigure}[b]{0.23\textwidth}
        \centering
        \includegraphics[width=\textwidth]{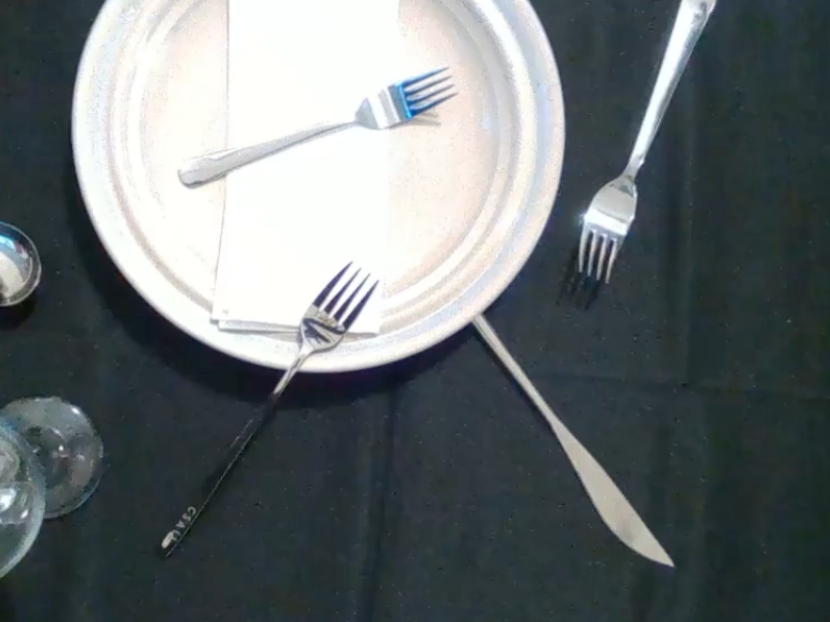}
        \caption{Input image}
        \label{fig:mean and std of net14}
    \end{subfigure}
    \hfill
    \begin{subfigure}[b]{0.23\textwidth}  
        \centering 
        \includegraphics[width=\textwidth]{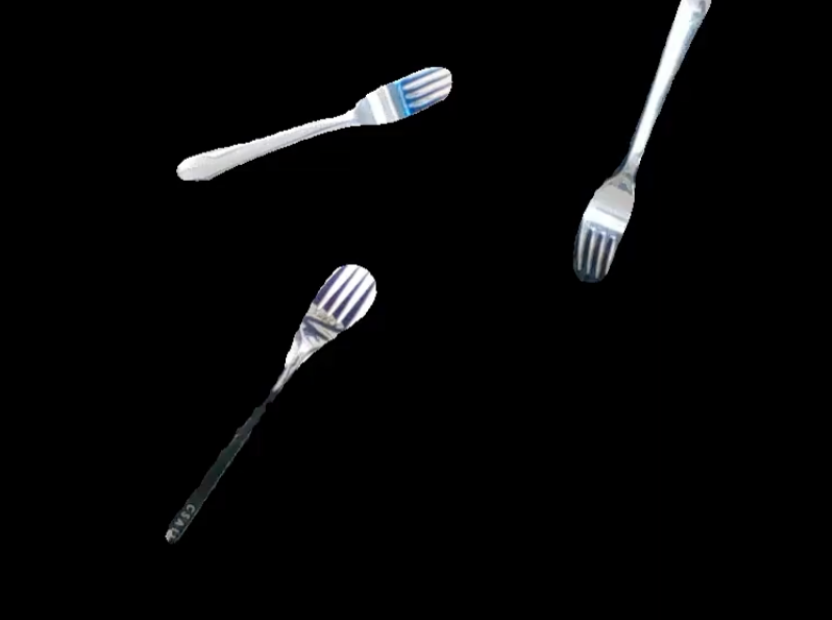}
        \caption{Segmented image}
        \label{fig:mean and std of net24}
    \end{subfigure}
    \vskip\baselineskip
    \begin{subfigure}[b]{0.23\textwidth}   
        \centering 
        \includegraphics[width=\textwidth]{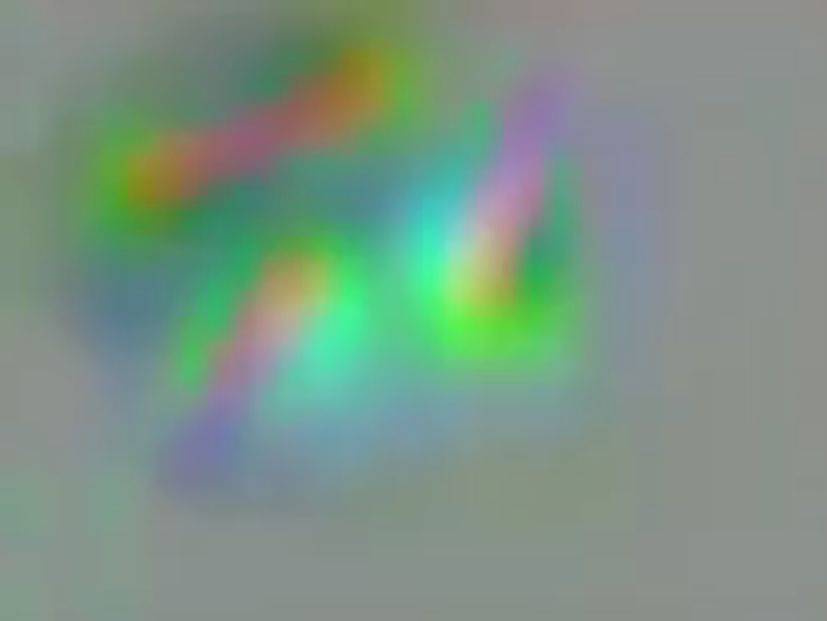}
        \caption{Output descriptors}  
        \label{fig:mean and std of net34}
    \end{subfigure}
    \hfill
    \begin{subfigure}[b]{0.23\textwidth}   
        \centering 
        \includegraphics[width=\textwidth]{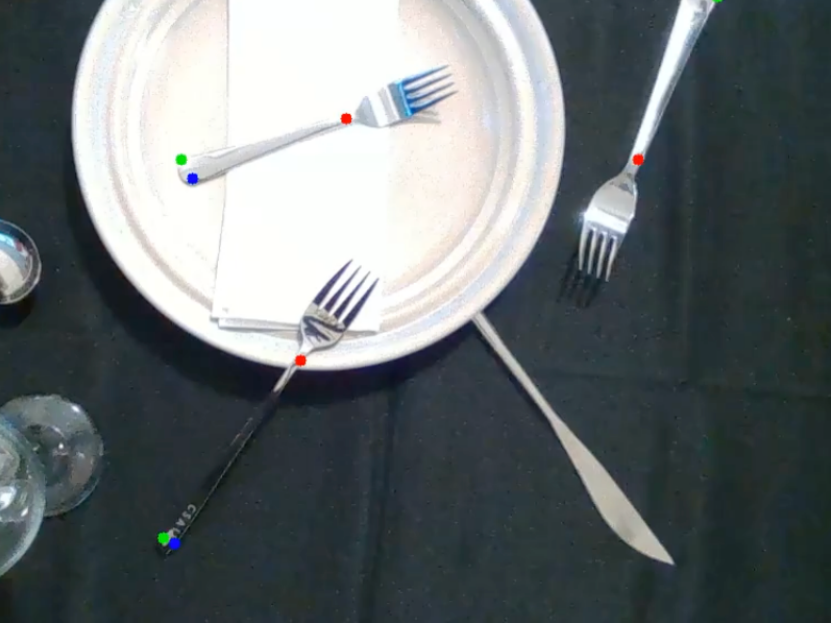}
        \caption{Tracked keypoints}
        \label{fig:mean and std of net44}
    \end{subfigure}
    }
    \caption{\textbf{Grasping pipeline.} We feed the input image (a) into a segmentation model to generate the segmented image (b), which is then taken as input to predict dense descriptors (c). We manually define a set of semantic keypoints (d) and track them using the descriptors. Finally, we perform triangulation on stereo image pairs to derive keypoints' 3D locations and the corresponding grasp pose. \textbf{Click  the  image  to  play  the  video  in  a  browser.}}
    \label{fig:grasping_pipeline}
\end{figure}

\begin{figure}[h!]
    \centering
    \href{https://youtu.be/_zN-wVwPH1s?t=273}{
        \includegraphics[width=0.48\textwidth]{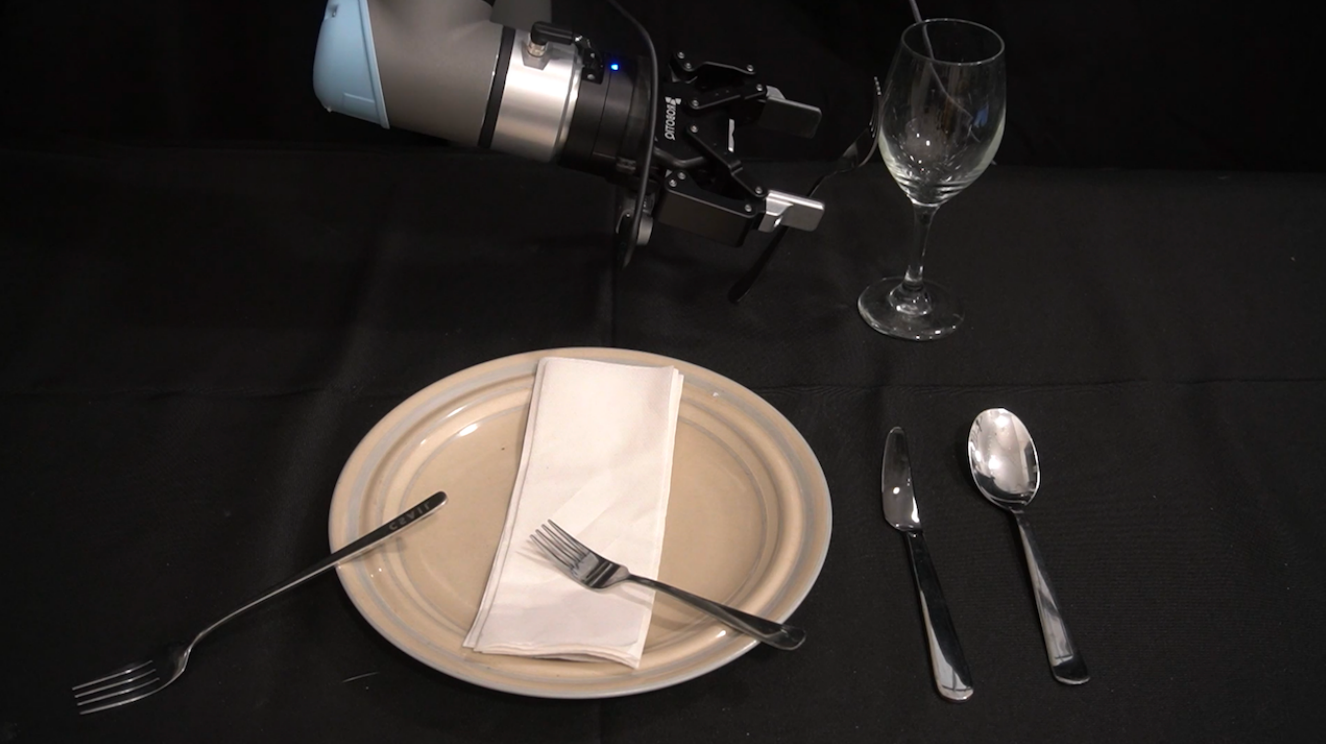}
    }
    \caption{\textbf{6-DoF pick and place.} We show that our robot can accurately grasp objects that are not in the training data with SE(3) grasp poses. \textbf{Click  the  image  to  play  the  video  in  a  browser.}
    }
    \vspace{-0.4em}
    \label{fig:demo}
\end{figure}

We use a UR5 robot with a Robotiq 2F-85 parallel jaw gripper. A RealSense D415 camera is mounted on the robot arm and precisely calibrated for both intrinsics and extrinsics. We illustrate the grasping pipeline in~\figref{fig:grasping_pipeline}, and we show the pick and place in action in~\figref{fig:demo}.
\end{appendix}

\bibliographystyle{IEEEtran}
\bibliography{IEEEexample}

\begin{thebibliography}{10}
\providecommand{\url}[1]{#1}
\csname url@samestyle\endcsname
\providecommand{\newblock}{\relax}
\providecommand{\bibinfo}[2]{#2}
\providecommand{\BIBentrySTDinterwordspacing}{\spaceskip=0pt\relax}
\providecommand{\BIBentryALTinterwordstretchfactor}{4}
\providecommand{\BIBentryALTinterwordspacing}{\spaceskip=\fontdimen2\font plus
\BIBentryALTinterwordstretchfactor\fontdimen3\font minus
  \fontdimen4\font\relax}
\providecommand{\BIBforeignlanguage}[2]{{%
\expandafter\ifx\csname l@#1\endcsname\relax
\typeout{** WARNING: IEEEtran.bst: No hyphenation pattern has been}%
\typeout{** loaded for the language `#1'. Using the pattern for}%
\typeout{** the default language instead.}%
\else
\language=\csname l@#1\endcsname
\fi
#2}}
\providecommand{\BIBdecl}{\relax}
\BIBdecl

\bibitem{mildenhall2020nerf}
B.~Mildenhall, P.~P. Srinivasan, M.~Tancik, J.~T. Barron, R.~Ramamoorthi, and
  R.~Ng, ``Nerf: Representing scenes as neural radiance fields for view
  synthesis,'' in \emph{{ECCV}}, 2020.

\bibitem{rublee2011orb}
E.~Rublee, V.~Rabaud, K.~Konolige, and G.~Bradski, ``Orb: An efficient
  alternative to sift or surf,'' in \emph{ICCV}, 2011.

\bibitem{lowe2004distinctive}
D.~G. Lowe, ``Distinctive image features from scale-invariant keypoints,''
  \emph{IJCV}, 2004.

\bibitem{bay2008speeded}
H.~Bay, A.~Ess, T.~Tuytelaars, and L.~Van~Gool, ``Speeded-up robust features
  ({SURF}),'' \emph{Computer vision and image understanding}, 2008.

\bibitem{detone2018superpoint}
D.~DeTone, T.~Malisiewicz, and A.~Rabinovich, ``Superpoint: Self-supervised
  interest point detection and description,'' in \emph{Computer Vision and
  Pattern Recognition Workshops}, 2018.

\bibitem{GLUNet_Truong_2020}
P.~Truong, M.~Danelljan, and R.~Timofte, ``{GLU-Net}: Global-local universal
  network for dense flow and correspondences,'' in \emph{CVPR}, 2020.

\bibitem{PDCNet_Truong_2021}
P.~Truong, M.~Danelljan, L.~V. Gool, and R.~Timofte, ``Learning accurate dense
  correspondences and when to trust them,'' in \emph{{CVPR}}, 2021.

\bibitem{jiang2021cotr}
W.~Jiang, E.~Trulls, J.~Hosang, A.~Tagliasacchi, and K.~M. Yi, ``{COTR:
  Correspondence Transformer for Matching Across Images},'' in \emph{ICCV},
  2021.

\bibitem{dosovitskiy2015flownet}
A.~Dosovitskiy, P.~Fischer, E.~Ilg, P.~Hausser, C.~Hazirbas, V.~Golkov, P.~Van
  Der~Smagt, D.~Cremers, and T.~Brox, ``Flownet: Learning optical flow with
  convolutional networks,'' in \emph{ICCV}, 2015.

\bibitem{mayer2016large}
N.~Mayer, E.~Ilg, P.~Hausser, P.~Fischer, D.~Cremers, A.~Dosovitskiy, and
  T.~Brox, ``A large dataset to train convolutional networks for disparity,
  optical flow, and scene flow estimation,'' in \emph{CVPR}, 2016.

\bibitem{Rocco17}
I.~Rocco, R.~Arandjelovi\'c, and J.~Sivic, ``Convolutional neural network
  architecture for geometric matching,'' in \emph{CVPR}, 2017.

\bibitem{GOCor_Truong_2020}
P.~Truong, M.~Danelljan, L.~V. Gool, and R.~Timofte, ``{GOCor}: Bringing
  globally optimized correspondence volumes into your neural network,'' in
  \emph{{NeurIPS}}, 2020.

\bibitem{florencemanuelli2018dense}
P.~R. Florence, L.~Manuelli, and R.~Tedrake, ``Dense object nets: Learning
  dense visual object descriptors by and for robotic manipulation,'' in
  \emph{Conference on Robot Learning}, 2018.

\bibitem{schoenberger2016sfm}
J.~L. Sch\"{o}nberger and J.-M. Frahm, ``Structure-from-motion revisited,'' in
  \emph{CVPR}, 2016.

\bibitem{manuelli2019kpam}
L.~Manuelli, W.~Gao, P.~Florence, and R.~Tedrake, ``kpam: Keypoint affordances
  for category-level robotic manipulation,'' \emph{arXiv preprint
  arXiv:1903.06684}, 2019.

\bibitem{schoenberger2016mvs}
J.~L. Sch\"{o}nberger, E.~Zheng, M.~Pollefeys, and J.-M. Frahm, ``Pixelwise
  view selection for unstructured multi-view stereo,'' in \emph{ECCV}, 2016.

\bibitem{martinbrualla2020nerfw}
R.~Martin-Brualla, N.~Radwan, M.~S.~M. Sajjadi, J.~T. Barron, A.~Dosovitskiy,
  and D.~Duckworth, ``Nerf in the wild: Neural radiance fields for
  unconstrained photo collections,'' in \emph{CVPR}, 2021.

\bibitem{park2021nerfies}
K.~Park, U.~Sinha, J.~T. Barron, S.~Bouaziz, D.~B. Goldman, S.~M. Seitz, and
  R.~Martin-Brualla, ``Nerfies: Deformable neural radiance fields,''
  \emph{ICCV}, 2021.

\bibitem{yen2020inerf}
L.~Yen-Chen, P.~Florence, J.~T. Barron, A.~Rodriguez, P.~Isola, and T.-Y. Lin,
  ``{iNeRF}: Inverting neural radiance fields for pose estimation,'' in
  \emph{IROS}, 2021.

\bibitem{SucarICCV2021}
E.~Sucar, S.~Liu, J.~Ortiz, and A.~Davison, ``{iMAP}: Implicit mapping and
  positioning in real-time,'' in \emph{ICCV}, 2021.

\bibitem{drebin1988volume}
R.~A. Drebin, L.~Carpenter, and P.~Hanrahan, ``Volume rendering,'' \emph{ACM
  Siggraph Computer Graphics}, 1988.

\bibitem{kangle2021dsnerf}
K.~Deng, A.~Liu, J.-Y. Zhu, and D.~Ramanan, ``Depth-supervised {NeRF}: Fewer
  views and faster training for free,'' \emph{arXiv preprint arXiv:2107.02791},
  2021.

\bibitem{schmidt2016self}
T.~Schmidt, R.~Newcombe, and D.~Fox, ``Self-supervised visual descriptor
  learning for dense correspondence,'' \emph{IEEE Robotics and Automation
  Letters}, 2016.

\bibitem{florence2019self}
P.~Florence, L.~Manuelli, and R.~Tedrake, ``Self-supervised correspondence in
  visuomotor policy learning,'' \emph{IEEE Robotics and Automation Letters},
  2019.

\bibitem{manuelli2020keypoints}
L.~Manuelli, Y.~Li, P.~Florence, and R.~Tedrake, ``Keypoints into the future:
  Self-supervised correspondence in model-based reinforcement learning,''
  \emph{arXiv preprint arXiv:2009.05085}, 2020.

\bibitem{sundaresan2020learning}
P.~Sundaresan, J.~Grannen, B.~Thananjeyan, A.~Balakrishna, M.~Laskey, K.~Stone,
  J.~E. Gonzalez, and K.~Goldberg, ``Learning rope manipulation policies using
  dense object descriptors trained on synthetic depth data,'' in \emph{ICRA},
  2020.

\bibitem{ganapathi2020learning}
A.~Ganapathi, P.~Sundaresan, B.~Thananjeyan, A.~Balakrishna, D.~Seita,
  J.~Grannen, M.~Hwang, R.~Hoque, J.~E. Gonzalez, N.~Jamali \emph{et~al.},
  ``Learning to smooth and fold real fabric using dense object descriptors
  trained on synthetic color images,'' \emph{arXiv:2003.12698}, 2020.

\bibitem{choy2016universal}
C.~B. Choy, J.~Y. Gwak, S.~Savarese, and M.~Chandraker, ``Universal
  correspondence network,'' in \emph{NeurIPS}, 2016.

\bibitem{tancik2020fourfeat}
M.~Tancik, P.~P. Srinivasan, B.~Mildenhall, S.~Fridovich-Keil, N.~Raghavan,
  U.~Singhal, R.~Ramamoorthi, J.~T. Barron, and R.~Ng, ``Fourier features let
  networks learn high frequency functions in low dimensional domains,''
  \emph{NeurIPS}, 2020.

\bibitem{kajiya84}
J.~T. Kajiya and B.~P.~V. Herzen, ``Ray tracing volume densities,''
  \emph{SIGGRAPH}, 1984.

\bibitem{max95}
N.~Max, ``Optical models for direct volume rendering,'' \emph{IEEE TVCG}, 1995.

\bibitem{trucco1998introductory}
E.~Trucco and A.~Verri, \emph{Introductory techniques for 3-D computer
  vision}.\hskip 1em plus 0.5em minus 0.4em\relax Prentice Hall Englewood
  Cliffs, 1998.

\bibitem{he2017mask}
K.~He, G.~Gkioxari, P.~Doll{\'a}r, and R.~Girshick, ``Mask r-cnn,'' in
  \emph{ICCV}, 2017.

\bibitem{schops2017multi}
T.~Schops, J.~L. Schonberger, S.~Galliani, T.~Sattler, K.~Schindler,
  M.~Pollefeys, and A.~Geiger, ``A multi-view stereo benchmark with
  high-resolution images and multi-camera videos,'' in \emph{CVPR}, 2017.

\end{thebibliography}

\end{document}